\documentclass[10pt,twocolumn,letterpaper]{article}

\usepackage[pagenumbers]{cvpr} 

\usepackage{graphicx}
\usepackage{amsmath}
\usepackage{amssymb}
\usepackage{booktabs}
\usepackage{multirow}

\usepackage[pagebackref,breaklinks,colorlinks]{hyperref}

\usepackage[capitalize]{cleveref}
\crefname{section}{Sec.}{Secs.}
\Crefname{section}{Section}{Sections}
\Crefname{table}{Table}{Tables}
\crefname{table}{Tab.}{Tabs.}

\author{Kai-Fu Yang$^{\#}$,~Cheng Cheng$^{\#}$,~Shi-Xuan Zhao,~Xian-Shi Zhang,~Yong-Jie Li\\
University of Electronic Science and Technology of China, Chengdu, China\\
{\tt\small{yangkf@uestc.edu.cn, ajjry486@163.com}} \\
{\tt\small{zhaosx@std.uestc.edu.cn, zhangxianshi@uestc.edu.cn, liyj@uestc.edu.cn}}
}

\begin{document}

\title{Learning to Adapt to Light}

\maketitle

\begin{abstract}
Light adaptation or brightness correction is a key step in improving the contrast and visual appeal of an image. There are multiple light-related tasks (for example, low-light enhancement and exposure correction) and previous studies have mainly investigated these tasks individually. However, it is interesting to consider whether these light-related tasks can be executed by a unified model, especially considering that our visual system adapts to external light in such way. In this study, we propose a biologically inspired method to handle light-related image-enhancement tasks with a unified network (called LA-Net). First, a frequency-based decomposition module is designed to decouple the common and characteristic sub-problems of light-related tasks into two pathways. Then, a new module is built inspired by biological visual adaptation to achieve unified light adaptation in the low-frequency pathway. In addition, noise suppression or detail enhancement is achieved effectively in the high-frequency pathway regardless of the light levels. Extensive experiments on three tasks---low-light enhancement, exposure correction, and tone mapping---demonstrate that the proposed method almost obtains state-of-the-art performance compared with recent methods designed for these individual tasks. 

\end{abstract}

\section{Introduction}
Images are often taken under varying lighting conditions, which usually results in unsatisfactory quality and affects further computer-vision tasks, such as object detection or recognition. Therefore, image-brightness correction is a necessary step for obtaining a good visual appearance or facilitating subsequent visual understanding. On the other hand, this task strongly connects with the fundamental function of the biological visual system, that is, light adaptation \cite{rieke2009challenges}, which helps us maintain stable visual perception by reliably adapt to diverse light conditions. 

\begin{figure}
  \centering
  \includegraphics[width=3.2in]{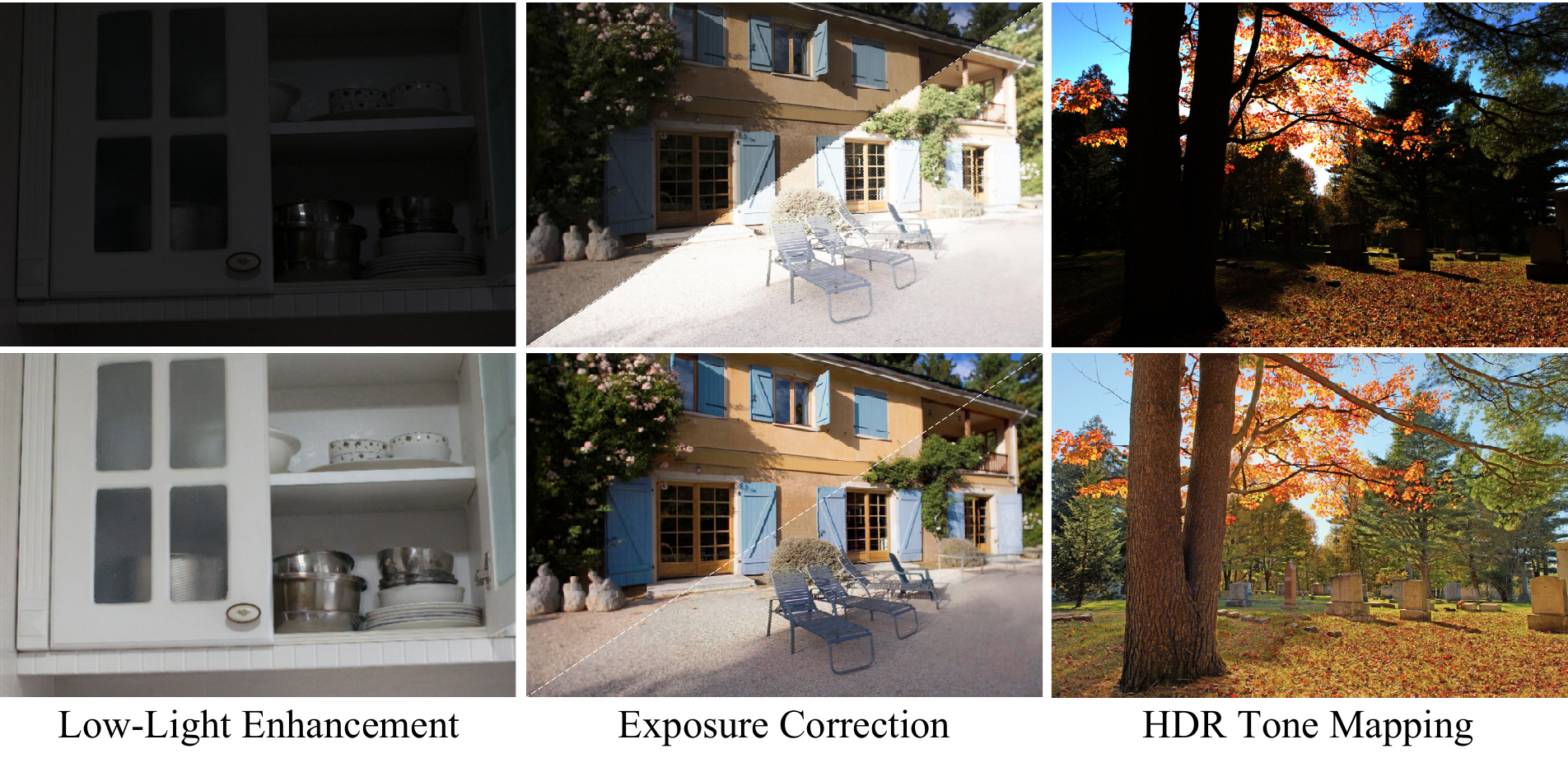}
  \caption{Examples of low-light enhancement, exposure correction, and HDR tone mapping. \textbf{Top}, input images; \textbf{Bottom}, results of enhancement. Note that the input HDR image is scaled linearly for better visibility.}
  \label{FigTasks}
\end{figure}

There are multiple tasks in computer vision that are aimed at achieving light adaptation, such as low-light enhancement \cite{guo2016lime, lore2017llnet, zhang2019kindling, zhang2021beyond}, exposure correction \cite{yu2018deepexposure, zhang2019dual, afifi2021learning}, and high-dynamic-range (HDR) tone mapping \cite{reinhard2010high, durand2002fast, liang2018hybrid, vinker2021unpaired}. Figure \ref{FigTasks} shows three examples of these enhancement tasks. The common key operator of these light-related tasks is to adjust the light level of the scene to an appropriate level and show more visual details. For example, low-light enhancement is aimed at improving the light level of dark regions to show more details and control noise at the same time. The exposure-correction task is also performed to adjust the exposure level (under- and overexposure) to show a clear image and enhance details. In addition, HDR tone mapping is aimed at compressing the dynamic range of HDR scenes to the low-dynamic-range (LDR) screen and preserving the details, which can also be treated as a kind of light adaptation for HDR inputs. 
 
However, most of the current methods are designed to cope with the aforementioned tasks individually, due to the different characteristics of these light-related tasks. For example, denoising is especially considered for low-light enhancement \cite{li2018structure, yang2019biological} because noises or artifacts are usually present in the dark regions. Nevertheless, HDR tone mapping may require a larger magnitude of light adaptation because of the HDR of input scenes, but less consideration of denoising \cite{reinhard2010high}. In addition, exposure correction must deal with both under- and overexposed scenes \cite{afifi2021learning}. 

It is interesting to consider whether these light-related tasks could be executed by a unified model, especially considering that such systems appear in the biological brain. Accordingly, this study aimed to build a unified network to handle multiple light-related enhancement tasks, including low-light enhancement, exposure correction, and HDR tone mapping. In addition, the mechanisms of two visual pathways \cite{schiller2010parallel} and visual adaptation in the early visual system \cite{rieke2009challenges} were considered to inspire the design of our network structure. 

To summarize the above, this work draws its inspiration directly from biological visual light adaptation, and the contributions of the proposed model are the following. (1) Frequency-based decomposition is used to separate the image-enhancement tasks into a common sub-problem of light adaptation and specific operators of noise suppression or detail enhancement for different tasks, instead of employing the widely used reflection-illumination decomposition under the Retinex assumption. (2) A bio-inspired module is built to achieve light adaptation for multiple light-related enhancement tasks in the low-frequency pathway. By mapping the input image into multiple channels with a group of learnable Naka-Rushton (NR) functions, the light adaptation is achieved by fusing multiple channels with local features. (3) In the high-frequency pathway, a simple residual-based sub-network is designed to handle both noise suppression and detail enhancement.

In experiments, we demonstrated that the proposed method achieves quite competitive performance compared with state-of-the-art methods on three light adaptation-related tasks with a unified network framework. Furthermore, the proposed method is computationally fast and requires a quite low memory footprint compared with recent top-rank methods, e.g., KinD++ \cite{zhang2021beyond}.

\section{Related Works}
A large body of methods for image enhancement exists. One type of important traditional method is the histogram-based method, including histogram equalization and its variants \cite{pizer1987adaptive, pisano1998contrast}, which usually enhances the visibility of an image by mapping the histogram. Another type of traditional method is the Retinex-based method, which assumes that the image can be decomposed into reflection and illumination according to the Retinex  theory \cite{land1977retinex}, e.g., single-scale Retinex (SSR) \cite{jobson1997properties} and multi-scale Retinex (MSR) \cite{jobson1997multiscale}. Accordingly, the following works attempt to optimize the estimation of illumination maps, such as NPE \cite{wang2013naturalness}, LIME \cite{guo2016lime}, and SRIE \cite{fu2016weighted}. 

Recently, deep-learning methods have been widely used for various image-enhancement tasks and exciting progress has been made. For low-light image enhancement, Lore et al. first proposed a deep network (called LLnet) for contrast enhancement and denoising \cite{lore2017llnet}. In addition, numerous researchers have attempted to build deep-learning networks based on the Retinex assumption \cite{land1977retinex}, which usually divides the image into two components \cite{wei2018deep, zhang2019kindling, zhang2021beyond, wang2019underexposed}.  Other deep-learning-based low-light enhancement methods use different learning strategies, including unsupervised learning-based \cite{jiang2021enlightengan}, zero-shot learning-based \cite{guo2020zero}, and semi-supervised learning-based methods \cite{yang2020fidelity}.  

In contrast to the task of low-light image enhancement, which is a kind of underexposure enhancement, Afifi et al. recently proposed a new method to correct the exposure errors for both under- and overexposed images with a single model \cite{afifi2021learning}. The exposure-correction task can also be treated as image light adaptation, but requires handling both under-and overexposed conditions simultaneously \cite{yuan2012automatic,yu2018deepexposure,zhang2019dual}.
 
In addition, tone mapping (TM) is another light-related task with HDR as input. TM operators are usually designed to compress the dynamic range of HDR images to the standard dynamic range while maintaining details and natural appearance. Traditional TM methods usually employ global or local operators to preserve image contrast \cite{reinhard2010high, durand2002fast, fattal2002gradient}. The exposure fusion method is also used to achieve HDR TM \cite{mertens2009exposure, yang2018adaptive}. Inspired by the local adaptation mechanism of the biological visual system, some researchers have built models for TM based on the Retinex theory \cite{meylan2006high, meylan2007model} or neural circuit in the retina \cite{zhang2020retina}. Recent methods aimed to achieve TM with a deep generative adversarial network have also been reported \cite{montulet2019deep, rana2019deep, panetta2021tmo, vinker2021unpaired}. 

In contrast to the majority of previous works that focus on the aforementioned tasks individually, we first isolate the common sub-problem (i.e., light adaptation) of these light-related image-enhancement tasks and handle it with a unified model inspired by the visual adaptation mechanisms in the biological visual system. In addition, noise suppression and detail enhancement are handled along another pathway. Hence, multiple light-related enhancement tasks are expected to be achieved with a unified framework.

\section{Method}
\subsection{Motivation}
\label{sec.mot}
\subsubsection{Frequency-based Decomposition}

Light-related visual-enhancement tasks contain the common sub-task of light adaptation, but also require different operators for noise or details. Thus, the first motivation of this study is to separate the common and specific sub-problems from multiple light-related enhancement tasks. In many previous studies, the image is divided into two components (reflection and illumination) following the Retinex theory and different types of degradation are handled in corresponding components \cite{wang2013naturalness,guo2016lime,fu2016weighted}. However, the Retinex assumption does not always hold and reflection-illumination decomposition is an ill-posed problem.  

In contrast, we adopt frequency-based decomposition, that is, decomposing images into low- and high-frequency pathways. The effectiveness of frequency-based decomposition in various enhancement-related tasks, such as nighttime defogging\cite{yan2020nighttime} and deraining \cite{fu2017removing}, has been demonstrated in previous works \cite{xu2020learning}. Furthermore, frequency-based decomposition is a biologically plausible approach based on the two visual pathways in the biological visual system \cite{yang2019biological}. Finally, noises and details are decomposed into the high-frequency pathway, thereby facilitating light adaptation in the low-frequency pathway and avoiding amplifying noises. In the high-frequency pathway, noise suppression or detail preservation may be easier to achieve regardless of the light level. 

\subsubsection{Bio-inspired Model for Unified Light Adaptation}
\label{NR.mov}
\begin{figure}
  \centering
  \includegraphics[width=3.3in]{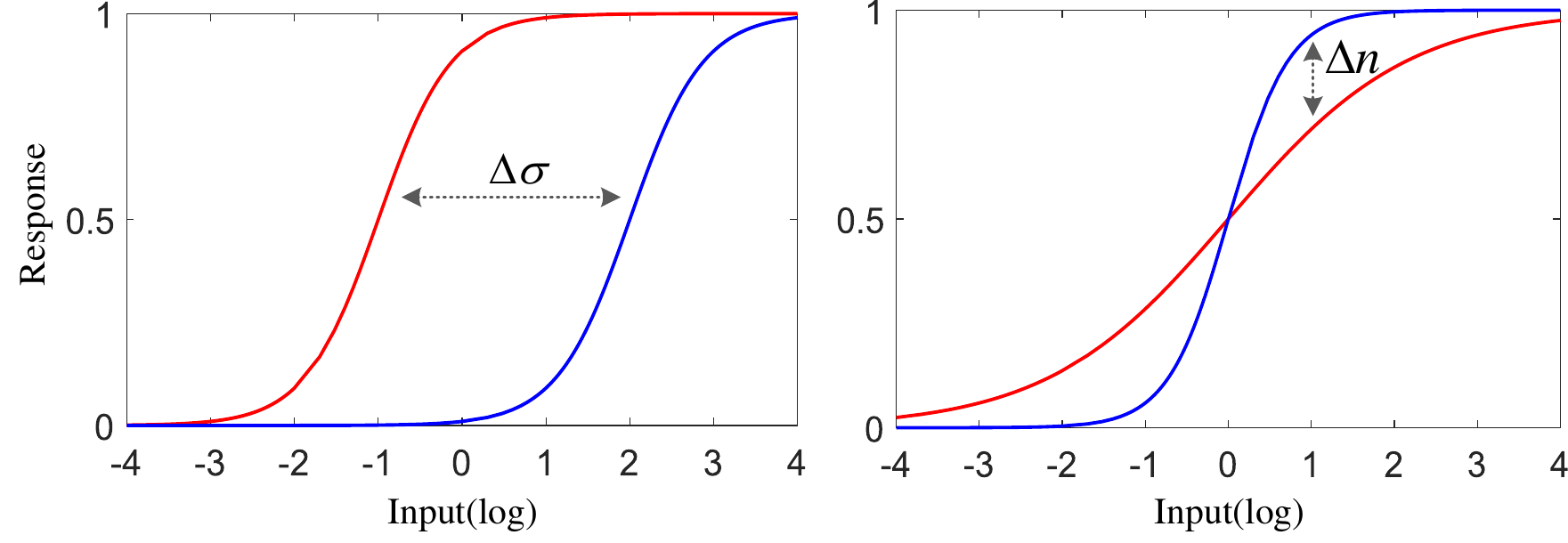}
  \caption{Characteristics of NR curves. $\sigma$ controls the mean of the S-shaped curve, i.e., light-adaptation level (\textbf{Left}), while $n$ is the scale of contrast adjustment by controlling the slope of the S-shaped curve (\textbf{Right}). }
  \label{FigNR}
\end{figure}

Light adaptation is an important mechanism in the biological visual system, which is used to keep the constancy of the perceptual level with varying visual scenes. The NR equation, which defines a kind of S-shaped response curve, is widely used to describe the process of visual light adaptation \cite{naka1966s}. The NR function can be expressed as 
\begin{equation}
 f(\sigma,n) = \frac{I^n}{ I^n + \sigma^n},
\end{equation}
where $I$ is the intensity of the visual input and $ \sigma $ is an adaptation factor used to control the mean value of the S-shaped curve. In biological visual systems, light adaptation is achieved by adjusting $ \sigma $ with varying lighting conditions. In addition, $ n $ is the scale of contrast adjustment by controlling the slope of the S-shaped curve. Figure \ref{FigNR} shows the basic characteristics of NR curves. Therefore, the NR functions provide a biologically plausible way to achieve light adaptation. However, how to select appropriate values of $ \sigma $ and $ n $ according to visual input or local change of lighting is a challenging problem.  

Inspired by the visual adaptation mechanisms in the biological visual system, we aimed to build a unified visual adaptation model by designing a learnable module to learn a group of adaptation parameters ($ \sigma $ and $ n $) for various scenes and regions. Thus, local light adaptation could be achieved by combining multiple channels with different NR curves according to local features. 

\subsubsection{Noise Suppression and Detail Enhancement}

After image decomposition, noises and details are usually present in the high-frequency pathway. To obtain clear images, noises should be removed or suppressed (e.g., in low-light enhancement), while details should be preserved or enhanced (e.g., HDR tone mapping). Therefore, noise suppression and detail enhancement could be achieved in the high-frequency pathway with a unified sub-network. Our model attempts to distinguish noises or details regardless of the light level, aiming to reduce the difficulty of fitting.

\subsection{Proposed Model}
\label{methods}
\begin{figure*}
  \centering
  \includegraphics[width=6.8in]{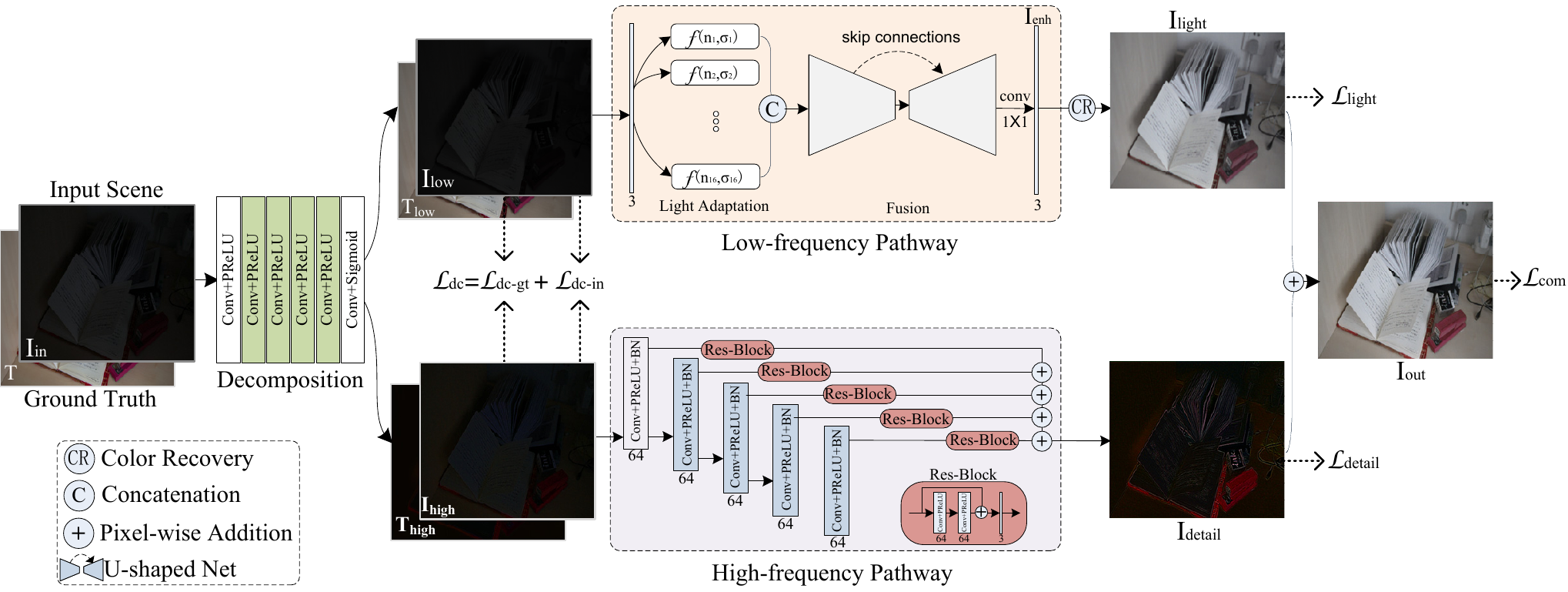}
  \caption{Flowchart of proposed network (LA-Net). The convolutional layers marked in the same colors in the decomposition network and the high-frequency pathway share the weights. More experiments about weight sharing can be found in the supplementary materials.}
  \label{FigFlowchart}
\end{figure*}

According to the description in Section \ref{sec.mot}, we propose a new network for image enhancement with the two-pathway and visual adaptation mechanisms. The pipeline of the proposed method is shown in Fig. \ref{FigFlowchart}. Specifically, the input image is first decomposed into low- and high-frequency components with a small convolutional network. Then, light adaptation is handled in the low-frequency pathway with a unified sub-network inspired by visual adaptation. Noise suppression and detail enhancement are achieved in the high-frequency pathway by introducing residual-based blocks that can prevent the disappearance of gradients, especially low values in the high-frequency pathway.

\subsubsection{Frequency-based Decomposition}

To decompose the input image into two pathways, we built a small convolutional sub-network that contains five Conv+PReLU layers and a Conv+Sigmoid layer. To achieve the decomposition, we employed total-variation (TV) loss, which has been widely used to decompose images into different frequency components \cite{aujol2006structure}. However, in our work, we integrated TV loss in the CNN network and trained with an end-to-end style. The TV-like loss used in this work includes three terms as follows:

\begin{equation}
\label{E2}
\begin{aligned}
 \mathcal{L}_{dc-in} &= \omega_1\cdot\parallel I_{in}-(I_{high}+I_{low})\parallel_{2}^{2} \\
    &+ \omega_2\cdot\parallel I_{in}-I_{low}\parallel_{2}^{2} +   \omega_3\cdot \bigtriangledown I_{low}
 \end{aligned},
\end{equation}
where $\omega_1=100$, $ \omega_2=2 $, and $ \omega_3=1 $, which are set experimentally (see the supplementary materials). The branches of output are denoted as $ I_{high} $ and $I_{low}$, where $ I_{high} $ contains main high-frequency components such as edges and noises, while $I_{low}$ contains the luminance and color information. 

In addition, to constrain the light adaptation and detail processing in the low- and high-frequency pathways, respectively, the ground-truth image (denoted as $T$) is also decomposed with the same network (shared weights) to generate low- and high-frequency components (denoted as $T_{low}$  and $ T_{high} $, respectively) of the ground-truth image. The loss function for the ground-truth image (denoted as $ \mathcal{L}_{dc-gt} $) is similar to Eq.(\ref{E2}), that is, 

\begin{equation}
\label{E3}
\begin{aligned}
 \mathcal{L}_{dc-gt} &= \omega_{1} \cdot\parallel T-(T_{high}+T_{low})\parallel_{2}^{2}\\
    &+\omega_{2} \cdot\parallel T -T_{low}\parallel_{2}^{2} + \omega'_{3} \cdot \bigtriangledown T_{low}
 \end{aligned},
\end{equation}
where we experimentally set $ \omega'_{3}= 5$. This is because the larger weight on the third term can balance the information between the high frequency component of low-light input and the ground-truth image, considering the general larger values of the low-frequency component in the ground-truth image (normal light). Finally, the final loss used for the decomposing network is

\begin{equation}
\label{E4}
\mathcal{L}_{dc} = \mathcal{L}_{dc-in} + \mathcal{L}_{dc-gt}.
\end{equation}

\subsubsection{Light-adaptation Model}

\begin{figure}
  \centering
  \includegraphics[width=3.2in]{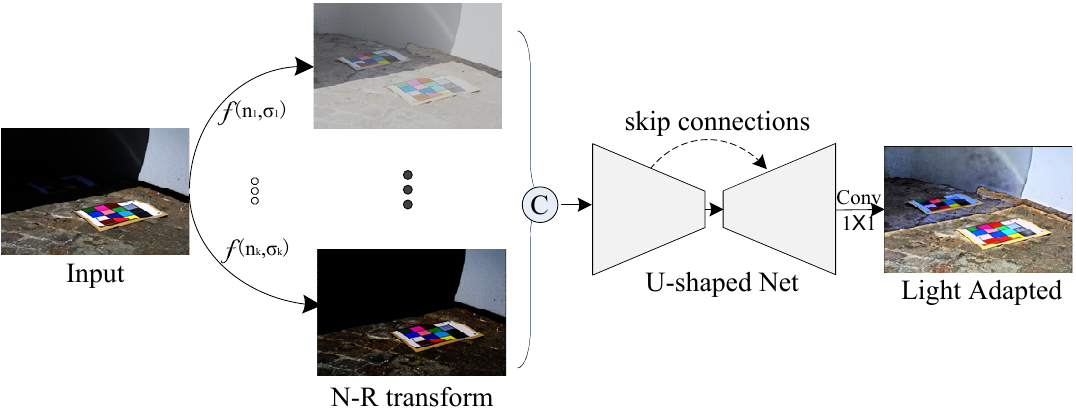}
  \caption{The computational flow of light-adaptation model with learnable NR functions. \textcircled{c} is a concatenation operator.}
  \label{FigNRnet}
\end{figure}

The key aspect of this work is to achieve light adaptation in the low-frequency pathway. Based on the NR function with the learnable parameters (i.e., $\sigma$ and $n$; see Section \ref{NR.mov}), we first map the input image into multiple channels to obtain images with different light levels. Thus, local light adaptation can be achieved by fusing the multiple channels according to local features. Figure \ref{FigNRnet} shows the computational flow of light adaptation model. The input image is firstly mapped into multiple channels with learnable NR functions. Each channel contains the information adapted to a specific light level, that is, specific regions are enhanced in each channel. Then, the outputs of all NR functions are concatenated and then fed into a small U-shaped net (3-layer U-Net with 32 output channels, presented in the supplementary materials) to integrate the light information. Finally, the output of the U-shaped net is fed through a $ 1\times1 $ convolutional layer to obtain the enhanced image.

With end-to-end learning, the proposed model can learn a group of NR functions (with learned $ \sigma $ and $n$) and express multiple light levels for various light-adaptation tasks. Meanwhile, the fusion net integrates the local light according to the learned features with convolutional layers. Finally, the light in the low-frequency component of the input image is corrected and light adaptation is achieved.

An additional step is used to recover the color of input scenes because light adjustment usually leads to color shifting. The enhanced image after light adaptation is denoted as $ I^{c}_{enh} $, and then the light-enhanced image with color recovery is obtained as 

\begin{equation}
\label{E5}
I^{c}_{light} = M_{enh} \cdot \frac{I^{c}_{enh}}{M_{low}},  c \in \{R,G,B\},
\end{equation}
where $ M_{low} $ and $ M_{enh} $ are the mean maps of $ I^{c}_{low} $ and $ I^{c}_{enh} $, respectively. This equation keeps the color of the output image the same as that of the input image regardless of light adaptation. Finally, a loss function is used to guide the learning of the light-adaptation in the low-frequency pathway, i.e., 
\begin{equation}
\label{E6}
\mathcal{L}_{light} = \parallel I_{light} - T_{low} \parallel_{2}^{2}.
\end{equation}

\subsubsection{Noise Suppression and Detail Enhancement}

To enhance details and suppress potential noises, we designed a sub-network to process the high-frequency information ($ I_{high} $). The structure of the sub-network is shown in Fig. \ref{FigFlowchart}. Specifically, the basic residual-block (i.e.,Res-Block) is used in the proposed sub-network, which is aimed at avoiding the possible vanishing of gradients in back-propagation, which, as usual, results in small pixel values in the images of details. A loss function used in the low-frequency pathway is defined as

\begin{equation}
\label{E7}
\mathcal{L}_{detail} = \parallel I_{detail} - T_{high} \parallel_{2}^{2}.
\end{equation}

The proposed network is targeted to achieve noise suppression for low-light enhancement tasks and realize detail enhancement or preservation when facing some noise-free input scenes, such as HDR scenes.

\subsubsection{Combining Two Pathways}
The final result is obtained by adding the light-adjusted image ($ I_{light} $) from the low-frequency pathway and the detail-enhanced image ($I_{detail}$) from the high-frequency pathway.
\begin{equation}
\label{E8}
I_{output} = I_{light}+I_{detail}.
\end{equation}

In addition, a $L_2$ loss function is used to polish the final results after combining the two pathways, that is, 
\begin{equation}
\label{E9}
\mathcal{L}_{com} = \parallel I_{output} - T \parallel_{2}^{2}.
\end{equation}

Finally, the widely used perceptual loss (denoted as $ \mathcal{L}_{Pce} $) is also additionally used to keep the constant in feature space encoded by the VGG16 network \cite{johnson2016perceptual}. Therefore, the total loss function is   
\begin{equation}
\label{E10}
\begin{aligned}
\mathcal{L} &= \lambda_1 \cdot \mathcal{L}_{dc}+ \lambda_2 \cdot \mathcal{L}_{light} + \lambda_3 \cdot \mathcal{L}_{detail} \\
&+ \lambda_4 \cdot \mathcal{L}_{com} +\lambda_5 \cdot \mathcal{L}_{Pce}
\end{aligned},
\end{equation}
where $ \lambda_1=\lambda_3=\lambda_5=1 $, $ \lambda_2=10 $, and $ \lambda_4=5 $. The analysis of main parameter settings can be found in the supplementary materials. 

\section{Experimental Results}
\subsection{Implementation Details}
The proposed network was trained on one NVIDIA Titan Xp graphical processing unit (GPU) running the Pytorch framework. The Adam optimizer was used to train the network. The initial learning rate for the decomposition network was set to 0.0002 and scaled by 0.5 each 50 epochs, after a total of 100 epochs. The learning rates of the sub-networks in low- and high-frequency pathways were set to 0.0001. The reason for different learning-rate strategies used for sub-networks is that decomposition is the priority at the start stage of training, while the training will focus on image enhancement when the decomposition reaches a certain degree. The weight decay was set to 0.0001 and the model was trained in a total of 200 epochs with a batch size of 2. All training images were resized to  $512\times512$. In particular, the initial parameters were $\sigma=0.5$ and $n\in[0.5, 8]$ with equal intervals for all NR functions.

To compress the size of the proposed model, the convolutional layers with the same structures in the decomposition network and the high-frequency pathway share the weights. More experiments and discussions about weight sharing can be found in the supplementary materials.

\subsection{Performance Evaluation}
\label{sec.eval}
The performance of the proposed model is evaluated on three light-related image-enhancement tasks, namely, (1) low-light enhancement (LLE), (2) exposure correction (EC), and (3) HDR tone mapping (TM). The main characteristics of each task are summarized in Table \ref{T1}, which shows that the common operator of all three tasks is light adaptation. In this subsection, we present the experimental results of each task accordingly. 

\begin{table*} 
  \centering
  \caption{Main characteristics of three light-related image-enhancement tasks.}
  \label{T1}
  \begin{tabular}{llll}
  \hline
  Task & Light & Noise	& Focus \\
  \hline
  LLE  & Darkness  & Strong & Lighting and denoising \\
  EC  & Both over- and underexposure errors & Weak & Light correction and detail enhancement \\
  TM  & High dynamic range & Weak & Dynamic range compression and detail preservation \\
  \hline
  \end{tabular}
\end{table*}

\textbf{Low-light enhancement} mainly focuses on lighting the darkness regions of a scene, which also usually suffers from noises and artifacts. In this experiment, the widely used LOL dataset \cite{wei2018deep} is employed to train and evaluate the proposed model. This dataset contains 485 pairs of low-/normal-light images for training and 15 low-light images for testing; all images were captured from real scenes \cite{wei2018deep}.

The existing LLE methods of KinD++ \cite{zhang2021beyond}, KinD \cite{zhang2019kindling}, Retinex-Net \cite{wei2018deep}, GLAD \cite{wang2018gladnet}, DRBN \cite{yang2020fidelity}, EnlightenGAN \cite{jiang2021enlightengan}, Zero-DCE \cite{guo2020zero}, and LIME \cite{guo2016lime} are used for comparison. Popular metrics, including PSNR, SSIM \cite{wang2004image}, and NIQE \cite{mittal2012making} are adopted for quantitative comparisons. PSNR and SSIM are reference metrics, while NIQE is a non-reference metric. Table \ref{T2} lists the numerical results of all compared methods on the test set of the LOL dataset, which shows that the proposed LA-Net obtains the highest PSNR and second-highest SSIM. Therefore, the proposed LA-Net achieves quite competitive performance compared with the recent state-of-the-art KinD++ method \cite{zhang2021beyond}. More experiments on other low-light datasets show the similar results, provided in the supplementary materials.

\begin{table}
  \centering
  \caption{Numerical results for the test set of LOL dataset (15 images). An asterisk (*) indicates that these results were reproduced with the public model from the authors, and other compared results were taken directly from the paper of Kind \cite{zhang2019kindling}and Kind++\cite{zhang2021beyond}. $\uparrow$ indicates higher values are better, while $\downarrow$ indicates lower values are better
perceptual quality.}
  \label{T2}
  \begin{tabular}{lccc}
  \hline
  Method & PSNR$\uparrow$ & SSIM$\uparrow$ & NIQE$\downarrow$\\
  \hline
  Zero-DCE* \cite{guo2020zero}  & 14.83	& 0.53 & 8.22\\
  Retinex-Net \cite{wei2018deep} & 16.77 & 0.56 & 8.89\\
  LIME \cite{guo2016lime} & 16.76 & 0.56 & 8.38\\
  EnlightenGAN* \cite{jiang2021enlightengan} &	17.37 &	0.63 & 4.89 \\
  DRBN* \cite{yang2020fidelity} &	18.78 &	\textbf{0.82} & 5.11\\
  GLAD \cite{wang2018gladnet}	& 19.80	& 0.65 & 6.48\\
  KinD \cite{zhang2019kindling} &	20.86 &	0.80 & 5.15\\
  KinD++ \cite{zhang2021beyond} &	21.30 &	\textbf{0.82} & 3.88\\
  LA-Net & \textbf{21.71} & 0.81 & \textbf{3.10}\\
  \hline
  \end{tabular}
\end{table}

\begin{figure*}
  \centering
  \includegraphics[width=6.8in]{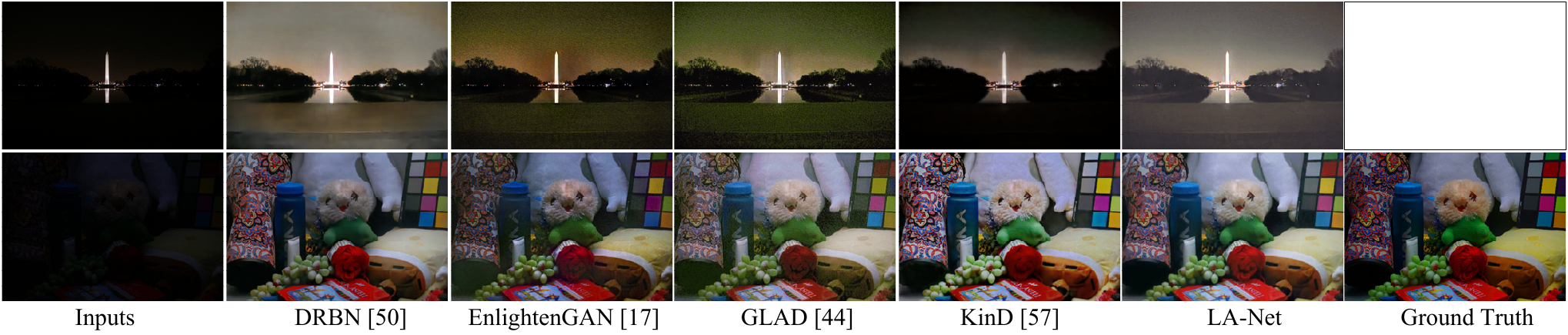}
  \caption{Visual comparisons of two low-light images.}
  \label{FigLLE}
\end{figure*}

Figure \ref{FigLLE} compares two low-light images. Results reveal the proposed method usually obtains better light and details in the dark regions. Meanwhile, noises are well suppressed, especially compared with GLAD \cite{wang2018gladnet} and DRBN \cite{yang2020fidelity}. In addition, the proposed method obtains similar or slightly better results compared with Kind \cite{zhang2019kindling}, which is consistent with the metrics listed in Table \ref{T2}.

\textbf{Exposure correction} focuses on correcting images with both over- and underexposure errors in real scenes. For this task, a recent large-scale image dataset is available in which images are rendered with a wide range of exposure errors and corresponding ground-truth images rendered manually by five photographers \cite{afifi2021learning} are provided. The dataset contains a total of 24,330 images, including 17,675 images in the training set, 750 images in the validation set, and 5,905 images in the test set. In this experiment, only 1,000 images and corresponding ground truths randomly selected from the training set are used to train the proposed network due to small-scale learnable parameters. More analysis on the influence of different numbers of training images can be found in the supplementary materials.

In addition, PSNR, SSIM and perceptual index (PI) \cite{afifi2021learning, blau20182018, ma2017learning} are adopted to quantitatively evaluate the pixel-wise accuracy, following previous work by Afifi et al. \cite{afifi2021learning}. The methods are evaluated on the combined over- and underexposed images (5,905 images). Table \ref{T3} lists the numerical results of the proposed method and compared methods, which shows that our method obtains the best performance with PSNR and  SSIM. It should be noted that the values of compared methods are directly adopted from \cite{afifi2021learning}. In Table \ref{T3}, the results of HDR CNN \cite{eilertsen2017hdr}, DPED\cite{ignatov2017dslr}, and DPE\cite{chen2018deep} indicate the best versions shown in \cite{afifi2021learning}, that is, HDR CNN w/PS, DPED (BlackBerry), and DPE (S-FiveK). 

\begin{table}
  \centering
  \caption{Numerical results on test set of combined over- and underexposed images (5,905 images) \cite{afifi2021learning}. These scores of compared methods were taken directly from Ref. \cite{afifi2021learning}.}
  \label{T3}
  \begin{tabular}{lccc}
  \hline
  Method & PSNR$\uparrow$ & SSIM$\uparrow$ & PI $\downarrow$ \\
  \hline
  Zero-DCE \cite{guo2020zero} & 12.598 & 0.549 & 2.865 \\
  RetinexNet \cite{wei2018deep} & 11.633	& 0.607 & 3.105 \\
  Deep UPE \cite{wang2019underexposed} & 14.247 & 0.640 & 2.405  \\
  HDR CNN \cite{eilertsen2017hdr} & 17.032 & 0.687 & 2.267\\
  DPED \cite{ignatov2017dslr} & 17.890 & 0.671 & 2.564 \\
  DPE \cite{chen2018deep} &	17.510 & 0.677 & 2.621 \\
  Afifi et al. \cite{afifi2021learning} & 19.377 & 0.731 & \textbf{2.247} \\
  LA-Net & \textbf{20.704} & \textbf{0.819}  & 2.353 \\
  \hline
  \end{tabular}
\end{table}

\begin{figure}
  \centering
  \includegraphics[width=3.0in]{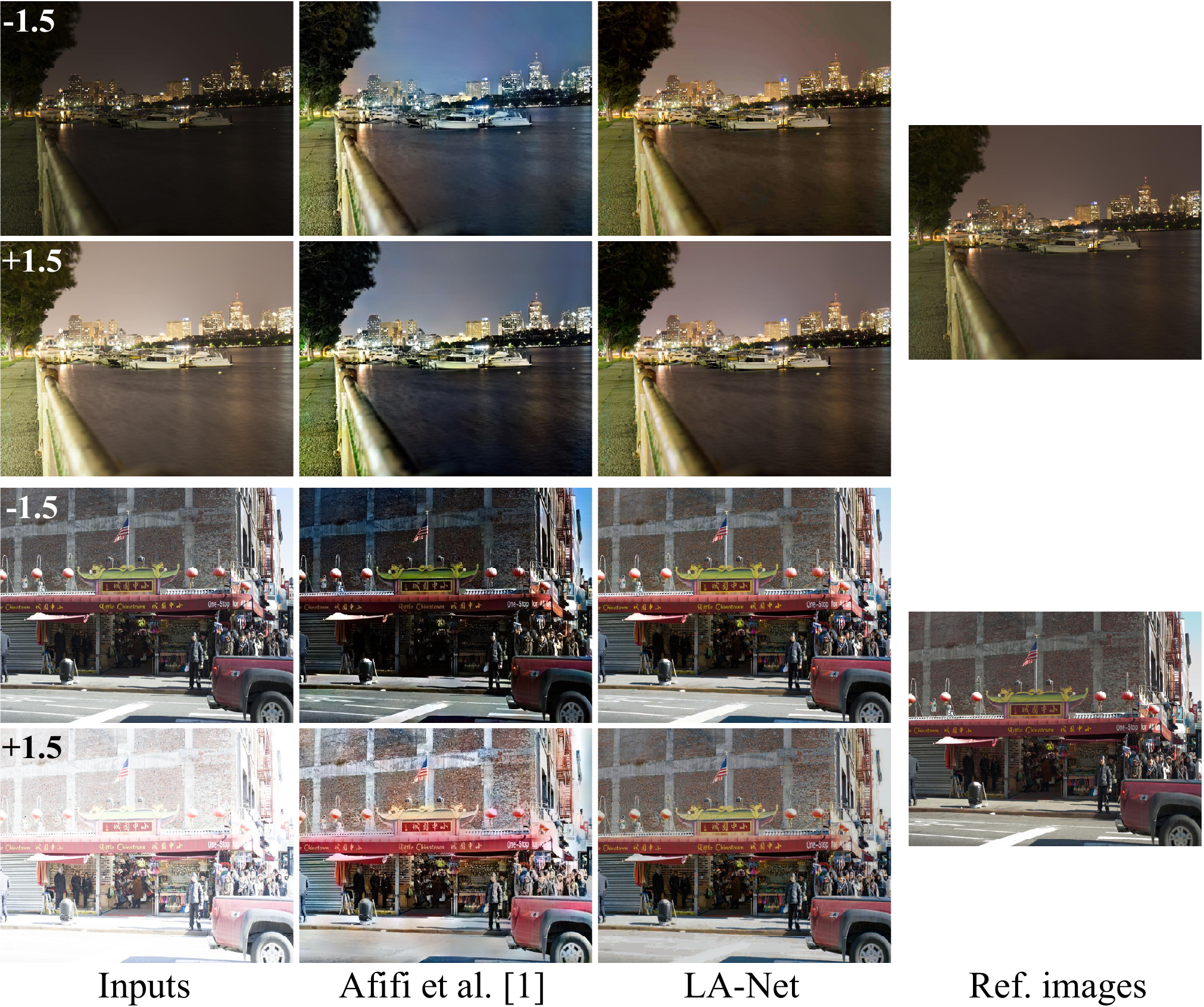}
  \caption{Comparisons of under- and overexposed images. Numbers at the top-left corners indicate the relative exposure values.}
  \label{FigEC}
\end{figure}

In addition, Fig. \ref{FigEC} presents comparisons of two scenes from the dataset used in Afifi et al. \cite{afifi2021learning}. The proposed method can adjust the over- and underexposed images with a single model, and obtains better results than the method proposed in Afifi et al. \cite{afifi2021learning}. For example, the color appearance of our results is more natural and closer to the reference images (see the first scene in Fig. \ref{FigEC}). In addition, the proposed method can recover the details in the underexposed regions better (see the second scene in Fig. \ref{FigEC}). 

\textbf{HDR tone mapping} is aimed at compressing the dynamic range of HDR scenes. Compared with LDR enhancement, HDR tone mapping requires larger dynamic range compression, but there is less concern regarding noises. In this experiment, the LVZ-HDR dataset \cite{panetta2021tmo} consisting of 456 images and 105 images from Internet \footnote{\url{http://www.hdrlabs.com/sibl/archive.html} (Creative Commons Attribution-Noncommercial-Share Alike 3.0 License)} are employed to train the proposed network. It should be noted that extra data augmentation applied due to the dynamic range of images in the LVZ-HDR dataset is limited. We augment the training data with $ I_{aug}=(I_{in}/max(I_{in}))^\beta $, where $\beta$ was uniformly selected at random between [0.7, 2.0] and used to control the dynamic range of HDR scenes. Finally, the model was evaluated on the HDR Photographic Survey (HDRPS) dataset \cite{fairchild2007hdr}, which contains 105 HDR images \footnote{\url{http://markfairchild.org/HDR.html}}. 

\begin{figure}
  \centering
  \includegraphics[width=3.2in]{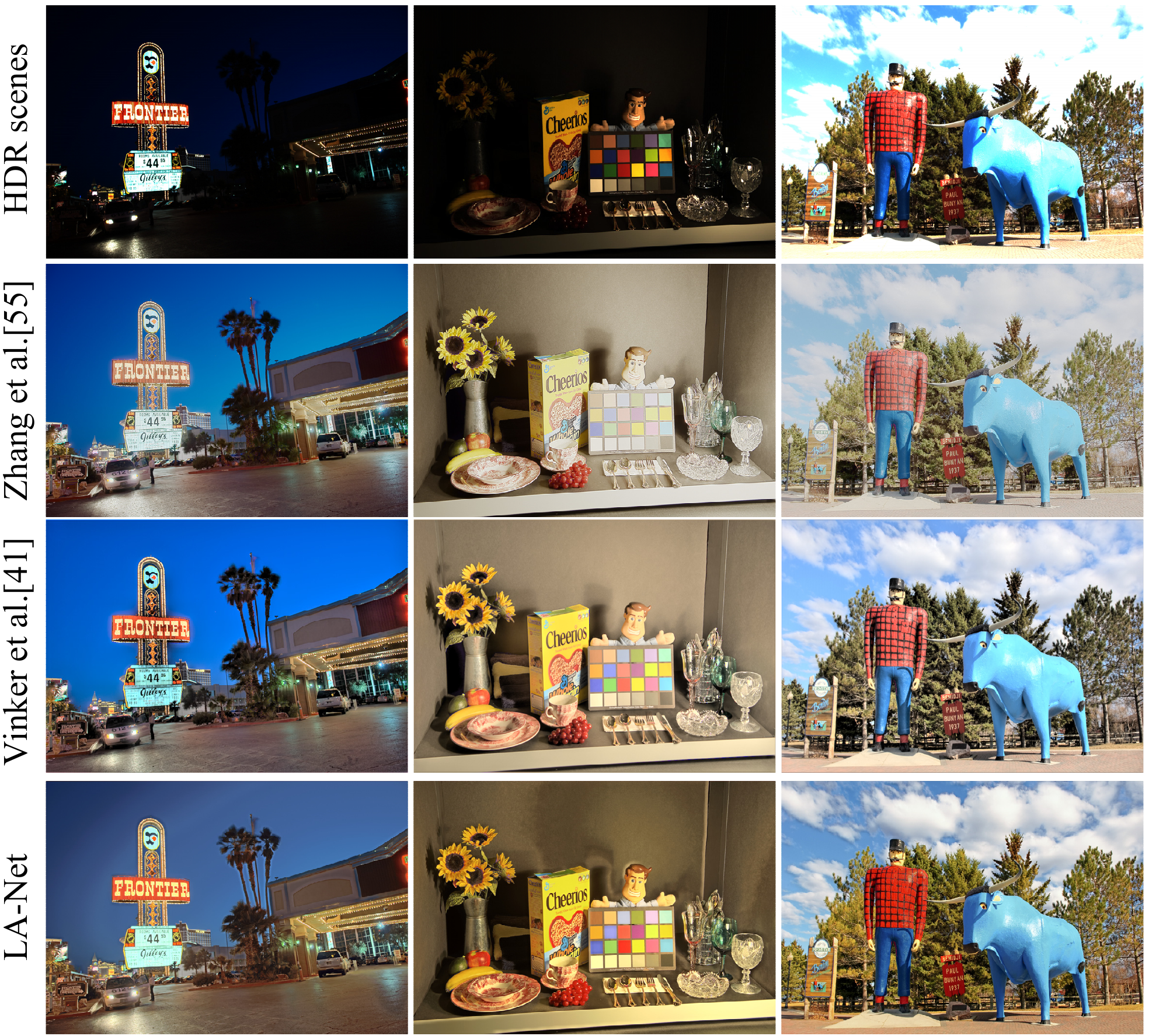}
  \caption{Visual comparisons of three HDR images. Note that the input HDR images are scaled linearly to clearly show the contents.}
  \label{FigHDR}
\end{figure}

In this experiment, multiple TM operators are used as compared methods, and MATLAB implementations of TMQI \cite{yeganeh2012objective} and BTMQI \cite{gu2016blind} are adopted to quantitatively evaluate performance. Table \ref{T4} lists the metrics obtained on the HDRPS dataset. Considering that the method of Vinker et al. \cite{vinker2021unpaired} outputs scaled images and image resizing affects the TMQI score \cite{cao2020adversarial}, we also list the TMQI and BTMQI scores with the same resizing of the result images, denoted as \textit{LA-Net(resized)}, for a fair comparison. Note that difference exists between our reproduced scores and the ones in the original paper of Vinker et al., which could be caused by the different implementations of TMQI and BTMQI. These results show that the proposed method achieves promising performance compared with considered methods. In addition, Fig. \ref{FigHDR} gives comparisons of several scenes from the HDRPS dataset. It can be seen that the results obtained by our method show the proper compression level and better color appearance, while the results obtained by Zhang et al. \cite{zhang2020retina} show overemphasized bright regions.

\begin{table}
  \centering
  \caption{Numerical results using the HDRPS dataset (105 images). An asterisk (*) denotes that these scores were taken directly from the respective paper. Two asterisks (**) denotes that these scores were reproduced with the public model from the authors.}
  \label{T4}
  \begin{tabular}{lcc}
  \hline
  Method & TMQI$\uparrow$ & BTMQI$\downarrow$\\
  \hline
  Liang et al. \cite{liang2018hybrid} & 0.8650 & 3.9710 \\ 
  Shibata et al. \cite{shibata2016gradient} & 0.877 & 3.4134 \\ 
  Zhang et al.* \cite{zhang2020retina} & 0.88 & 3.76 \\ 
  Rana et al.* \cite{rana2019deep} & 0.88 & -- \\
  ETMO* \cite{su2021explorable} & 0.8652 & -- \\
  Vinker et al.** \cite{vinker2021unpaired} & 0.8861 & 3.6447 \\ 
  \hline
  LA-Net & \textbf{0.8803} & \textbf{3.1728} \\
  LA-Net(resized) & \textbf{0.8901} & \textbf{3.1904} \\
  \hline
  \end{tabular}
\end{table}

\begin{figure}
  \centering
  \includegraphics[width=3.0in]{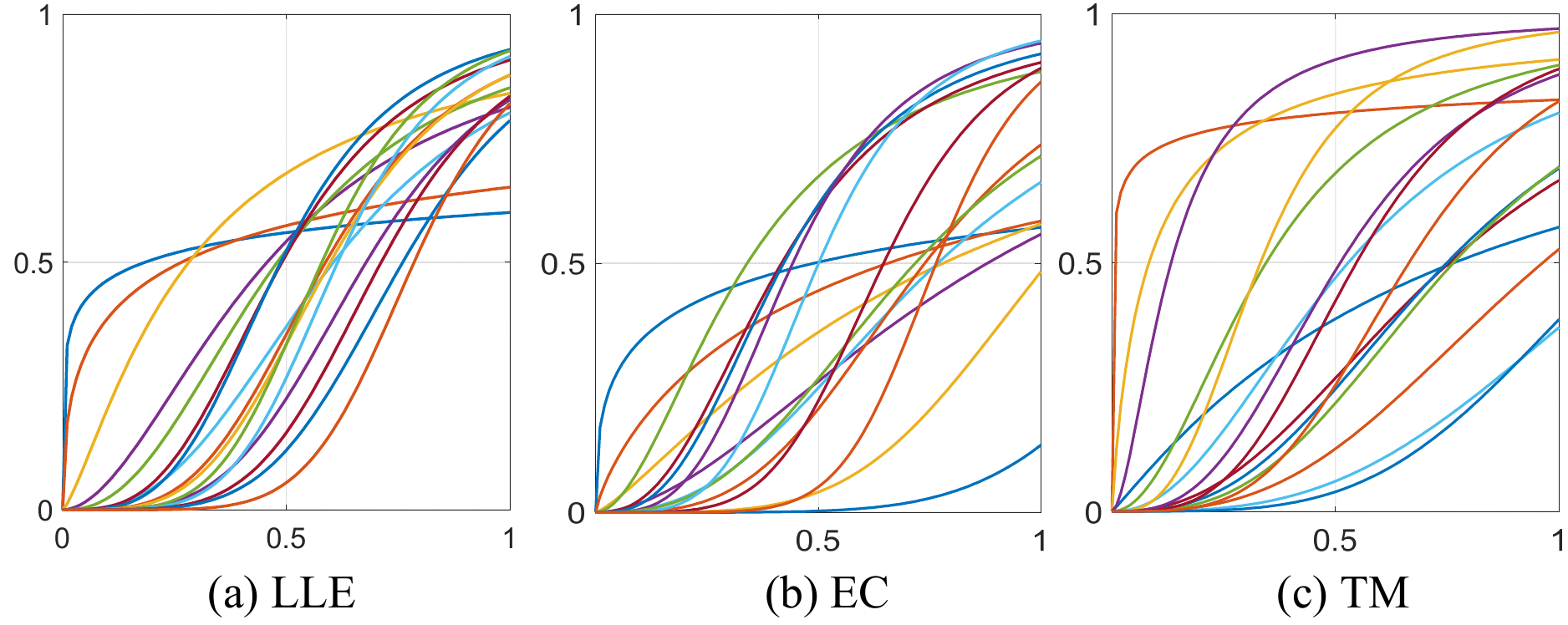}
  \caption{Learned NR curves for three tasks under study. Herein, NR curves with learned parameters are plotted on the linear axis to conveniently show the input-output relationship.}
  \label{FigNRresults}
\end{figure}

To further analyze the characteristics of the proposed light-adaptation module, the learned parameters of NR curves for the three tasks under study are shown in Fig. \ref{FigNRresults}. It can be seen from this figure that the optimal NR curves have a diverse distribution, so that the model can adapt to different light conditions. Generally, more learned NR curves contribute to lighting dark regions for LLE, shown in Fig. \ref{FigNRresults}(a), while the more diverse distribution for EC is due to the existing of both over- and underexposure errors (Fig. \ref{FigNRresults}(b)). In addition, learned NR curves have the most diverse distribution for TM because of the high dynamic range of inputs (Fig. \ref{FigNRresults}(c)). The ablation study on numbers of NR curves is listed in Section \ref{sec.AS} and corresponding learned NR curves can be found in the supplementary materials.
 
\subsection{Ablation Study and Parameter Analysis}
\label{sec.AS}
To demonstrate the contribution of frequency-based decomposition, we built a one-pathway model that inputs the original image into the sub-network of light adaptation (i.e., low-frequency pathway) and discarded the processing in the high-frequency pathway. The one-pathway model was also evaluated on three light-enhancement tasks with corresponding test sets, i.e., the LOL test dataset for LLE, Afifi et al.'s dataset for EC and the HDRPS dataset for TM. 

Table \ref{T5} lists the results of the one- and two-pathway models (the latter is the proposed LA-Net). On LLE and EC tasks, LA-Net significantly outperforms the one-pathway model, which benefits from the high-frequency pathway with noise suppression and detail enhancement. However, The one-pathway model also achieves slightly better results with TM tasks and outperforms the recent method of Vinker et al. \cite{vinker2021unpaired}. This is reasonable considering that the TM task mainly focuses on dynamic range compression and the input HDR scenes contain fairly weak noises. 

\begin{table}
  \centering
  \caption{Results of one- and two-pathway models (i.e., LA-Net).}
  \label{T5}
  \begin{tabular}{ccccc}
  \hline
  Method & Metric & LLE	& EC & TM  \\
  \hline
  \multirow{2}{*}{One-pathway} & PSNR & 15.570 & 17.495 & 0.8975\\
  \cline{2-4}
   & SSIM & 0.377 & 0.745 & (TMQI) \\
  \hline
  Two-pathway &	PSNR & 21.71 & 20.704 & 0.8803 \\
  \cline{2-4}
  (LA-Net) & SSIM & 0.805 & 0.819 & (TMQI) \\
  \hline
  \end{tabular}
\end{table}

\begin{figure}
  \centering
  \includegraphics[width=3in]{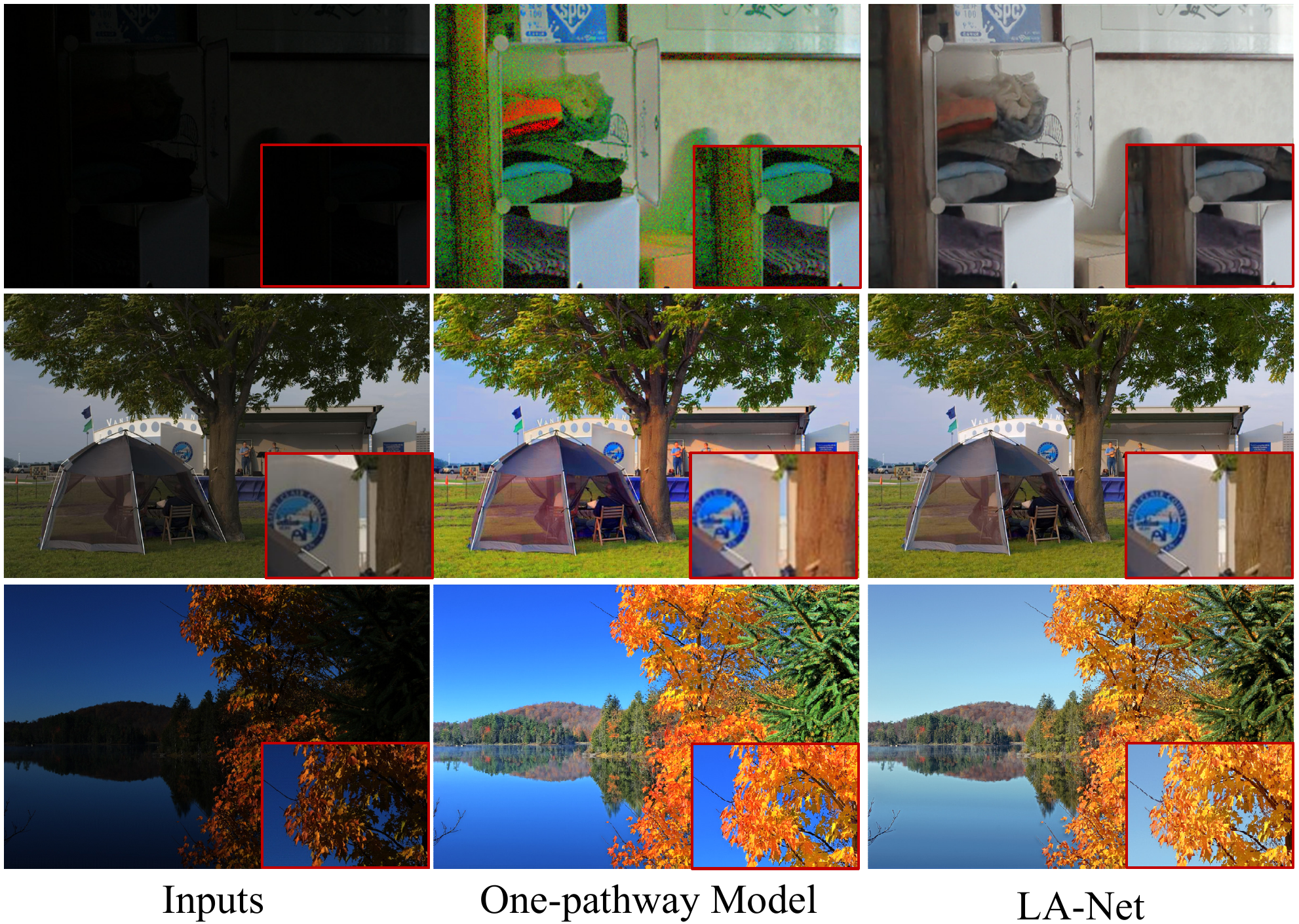}
  \caption{Visual comparisons of one- and two-pathway models on three tasks studied.}
  \label{FigOnTwo}
\end{figure}

Some visual comparisons are shown in Fig. \ref{FigOnTwo}. With the proposed light-adaptation model, both models can enhance the light well on the three tasks considered. However, LA-Net can well suppress noises in low-light images and enhance the details in images with exposure errors, but has little influence on the details of HDR scenes.

In addition, we further tested the influence of different numbers of NR curves on the LLE task with the LOL test dataset. Table \ref{T6} lists the results and runtimes averaged over the LOL test set (on a GPU) when different numbers of NR curves are used in the light-adaptation model (in the low-frequency pathway). Results show that the proposed model obtains acceptable performance when more than four NR curves are used, and the model is relatively robust with the number of NR curves. The corresponding learned NR curves can be found in the supplementary materials. 

\begin{table}
  \centering
  \caption{Ablation study of numbers of NR curves on the LLE task with the LOL test dataset containing 15 images.}
  \label{T6}
  \begin{tabular}{cccc}
  \hline
  \#(N-R curves) & PSNR &  SSIM & Time (ms) \\
  \hline
  4	 & 21.575 & 0.805 & 30 \\
  8	 & 21.653 & 0.803 & 40 \\
  12 & 21.350 & 0.802 & 55 \\
  16 & 21.713 & 0.805 & 60 \\
  32 & 21.580 & 0.802 & 110 \\
  \hline
  \end{tabular}
\end{table}

In addition, the proposed model has only 0.575M trainable parameters, which mainly benefits from the weight sharing strategy in the designing of network. By comparison, the recent top-rank method (KinD++ \cite{zhang2021beyond}) has more than 8.0M trainable parameters. Note that, additional experiments show that the parameters contained in the loss functions usually affect the final results, which is illustrated in the supplementary materials.

\section{Conclusions and Limitations}
This work proposes a new network inspired by multi-pathway processing and visual adaptation mechanisms in the biological visual system. In particular, a new light-adaptation module is proposed to handle the common sub-problem in light-related enhancement tasks. Experimental results show the proposed method achieves state-of-the-art performance on three enhancement tasks. 

Our method does have limitations. For the results in section \ref{sec.eval}, our method achieves good performance on light adaptation but with certain loss of contrast. As a future work, we plan to build a unified model to tackle more visual enhancement tasks by integrating light adaptation, contrast enhancement, and color correction mechanisms.

\section*{Acknowledgements}
This work was supported by the National Natural Science Foundation of China under Grant 62076055.

{\small
\bibliographystyle{ieee_fullname}
\bibliography{Refs}
}

\end{document}